\documentclass[runningheads]{llncs}
\usepackage[T1]{fontenc}
\usepackage{graphicx}

\usepackage[utf8]{inputenc} %
\usepackage[T1]{fontenc}    %
\usepackage{url}            %
\usepackage{booktabs}       %
\usepackage{amsfonts}       %
\usepackage{nicefrac}       %
\usepackage{microtype}      %
\usepackage{amsmath}
\usepackage[colorlinks=true, allcolors=blue]{hyperref}
\usepackage[dvipsnames,table,xcdraw]{xcolor}
\usepackage{xspace}
\usepackage{multicol}
\usepackage{multirow}
\usepackage{booktabs} %
\usepackage{makecell} %
\usepackage{wrapfig}
\usepackage{algorithm}
\usepackage{algpseudocode}
\usepackage{caption}

\usepackage{tcolorbox}
\usepackage{tabularx}
\usepackage{array}
\usepackage{placeins}
\tcbuselibrary{skins, breakable}

\newcommand{\data}{COCOLogic-V2\xspace}
\newcommand{\dataf}{\data-FS\xspace}

\usepackage{listings}
\definecolor{lightgray}{gray}{0.95}
\tcbuselibrary{listings, breakable, skins}

\tcbuselibrary{listings, breakable, skins}
\newtcblisting{promptbox}[1]{
  breakable,
  listing engine=listings,    %
  listing only,               %
  before skip=14pt plus 2pt,
  after skip=14pt plus 2pt,
  enhanced,
  sharp corners,
  colback=gray!5,
  colframe=gray!80,
  fonttitle=\bfseries\ttfamily\small,
  title=#1,
  attach boxed title to top left={yshift=-2mm, xshift=2mm},
  boxed title style={
    sharp corners, 
    colback=gray!80, 
    colframe=gray!80
  },
  listing options={
    basicstyle=\scriptsize\ttfamily,
    breaklines=true,
    columns=fullflexible,
    literate={_}{\textunderscore}1,
    basewidth=0.5em,
    lineskip=1pt
  },
  left=2mm,
  right=2mm,
  top=3mm, 
}

\usepackage{todonotes}
\newcounter{mycomment}

\usepackage{tikz}
\usetikzlibrary{
  positioning,
  arrows.meta,
  fit,
  backgrounds,
  calc,
  decorations.pathreplacing
}

\colorlet{ColI}{blue!22}          %
\colorlet{ColII}{teal!30}         %
\colorlet{ColIII}{violet!22}      %
\colorlet{ColNB}{orange!18}       %
\colorlet{ColFlip}{red!78!black}  %

\begin{document}
\title{\data: Identifying Logical Inconsistencies \\via Truly Hard-Negatives}

\author{David Steinmann\inst{1, 2} \and
Antonia Wüst\inst{1} \and
\\Kristian Kersting\inst{1, 2, 3}
\and Wolfgang Stammer\inst{4, 5}}
\authorrunning{D. Steinmann et al.}
\institute{Computer Science Department, TU Darmstadt \and
Hessian Center for AI (hessian.AI) \and
German Research Center for AI (DFKI) \and
Max Planck Institute for Informatics, SIC \and 
RTG Neuroexplicit Models\\
\email{david.steinmann@tu-darmstadt.de}}
\maketitle              %

\begin{abstract}
While interpretable models such as concept bottleneck models (CBMs) and program synthesis methods enable verification of model decisions, their evaluation is typically limited to simple tasks, leaving complex reasoning on real-world images largely unexplored. We introduce \data, an object-centric dataset for visual inductive reasoning on real-world images covering a broad subset of first-order logic. By categorizing samples into positive variants, near-boundary (NB), and far-from-boundary (FB) negatives, \data enables fine-grained diagnosis of model accountability. Our evaluations show that models tend to separate positive and FB samples well but fail on NB samples, while perceptual noise and large rule-induced search spaces pose additional challenges in few-shot settings. Together, these results highlight that visual inductive reasoning remains an open challenge and \data provides a concrete foundation for advancing methods in this direction.\footnote{Code and dataset: \url{https://github.com/ml-research/COCOLogic-V2}}

\end{abstract}

\setlength{\tabcolsep}{5pt}
\renewcommand{\figureautorefname}{Fig.\xspace}
\renewcommand{\tableautorefname}{Tab.\xspace}
\renewcommand{\sectionautorefname}{Sec.\xspace}
\renewcommand{\equationautorefname}{Eq.\xspace}
\renewcommand{\appendixautorefname}{Suppl.\xspace}
\renewcommand{\subsectionautorefname}{Sec.\xspace}
\renewcommand{\subsubsectionautorefname}{Sec.\xspace}
\newcommand{\algorithmautorefname}{Alg.\xspace}

\section{Introduction}
Transparency and accountability are crucial requirements for machine learning in high-stakes domains such as healthcare or finance \cite{Rudin19}. Interpretable models, such as concept bottleneck models (CBMs) \cite{koh2020concept,StammerSK21,espinosa2022concept,knab2026bottle} and neuro-symbolic programs \cite{mao2019neuro,VisProg,kamali2025neptune,shindo2023alpha,wust2026synthesizing}, address this by enabling manual verification of model decisions. However, their evaluation is typically limited to narrow settings: CBMs are often assessed on single-label image classification \cite{koh2020concept,espinosa2022concept,panousis2024coarse,shang2024incremental}, while program synthesis and neuro-symbolic approaches rely on symbolic representations for text editing \cite{ellis2023dreamcoder}, abstract pattern matching \cite{chollet2019measure}, rather than inductive reasoning \cite{mao2019neuro,VisProg,kamali2025neptune}. It therefore remains unclear to what extent such models can tackle more complex tasks, such as visual inductive reasoning on real-world data.

The recently introduced COCOLogic dataset \cite{steinmann2026object} takes a step in this direction, providing a reasoning-inspired classification task on real-world images from MSCOCO \cite{LinMBHPRDZ14}. However, its single-label framing introduces shortcuts; for example, a bowl alone predicts the "Unlikely Breakfast Guests" class, and the resulting class imbalance further obscures the logical task difficulty.
\begin{wrapfigure}{tr}{0.35\linewidth}
\includegraphics[width=\linewidth]{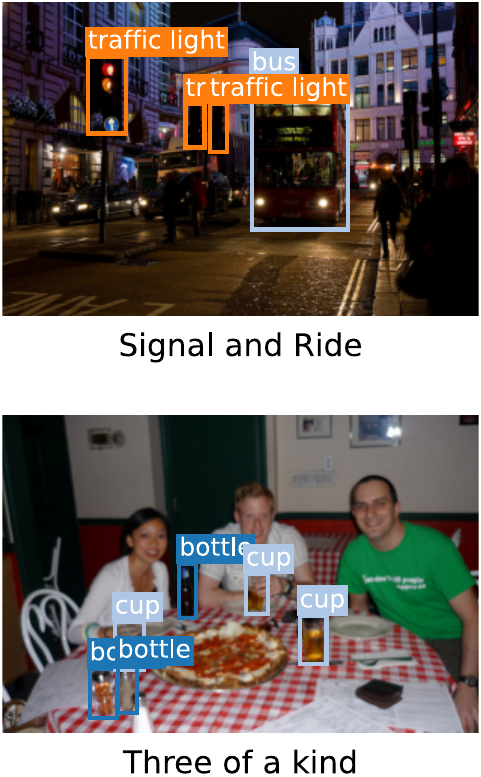}
\caption{\data requires reasoning over real-world images, where each logic rule depends on specific objects within an image, illustrated here for two example rules: \textit{Signal and Ride} and \textit{Three of a Kind}.}
\label{fig:intro}
\vspace{-0.7cm}
\end{wrapfigure}
To address these limitations, we introduce \data. Beyond reframing the task as multilabel classification to reduce shortcuts and class imbalance, \data demands reasoning over a wider range of first-order logical operations, including object counting and counting comparisons. Samples are categorized per rule into \textit{positive variants} (the different ways a rule can be fulfilled), difficult \textit{near-boundary} (NB) negatives close to the decision boundary, and \textit{far-from-boundary} (FB) negatives that can often be classified correctly without proper rule understanding, serving together as an automatic diagnostic tool for model accountability. We additionally provide \dataf, a curated small version for few-shot and in-context learning.

We evaluate several families of approaches on both dataset versions. On \data, CBM approaches and a black-box baseline reliably separate positive and FB samples but consistently struggle on NB samples, suggesting they exploit statistical patterns rather than learning the underlying rules. On \dataf, perceptual noise and large rule-induced search spaces pose additional challenges for in-context learning with VLMs and program synthesis approaches.
Altogether, this work makes the following contributions: (1) we introduce \data, an object-centric dataset for visual inductive reasoning over natural images; (2) we provide \dataf, a curated few-shot version for data-constrained settings; (3) we show that current interpretable approaches fall short, highlighting open challenges in this setting.

The remainder of the paper is structured as follows: In \autoref{sec:dataset}, we introduce \data and differentiate it from existing work. Afterward, we provide model evaluations in \autoref{sec:eval}, a short discussion (\autoref{sec:discussion}), and a conclusion in \autoref{sec:conclusion}.

\section{The \data Dataset}
\label{sec:dataset}

\begin{table}[t]
\centering
\caption{\textbf{\data rule definitions.} Rules are based on the occurrence or absence of object categories within an image. Some rules additionally require a specific number of certain objects to be present or compare the number of objects from different categories.}
\label{tab:rules}
\resizebox{\textwidth}{!}{
\begin{tabular}{l@{\hspace{0.5cm}}l}
\toprule
\textbf{Name} & \textbf{Logical Rule}\\
\midrule
Signal and Ride & \texttt{traffic light} $\wedge$ one of \{\texttt{bicycle}, \texttt{bus}, \texttt{train}\} \\
\midrule
Double Serving & Exactly two categories of \{\texttt{bottle}, \texttt{cup}, \texttt{pizza}\}\\
\midrule
Herd Alone & At least two objects of the same category of \{\texttt{cow}, \texttt{elephant}, \texttt{sheep}\} $\wedge$ no \texttt{person} \\
\midrule
Either Dog or Car & Either \texttt{dog} or \texttt{car} \\
\midrule
Three of a Kind & Exactly three \texttt{bowl} $\lor$ exactly three \texttt{cup} \\
\midrule
Car Majority & More \texttt{car} than \texttt{truck} $\wedge$ at least one of each \\
\midrule
Empty Seat & (\texttt{couch} $\lor$ \texttt{chair}) $\wedge$ no \texttt{person} \\
\midrule
Single Mode Traffic & Exactly one category of \{\texttt{bicycle, motorcycle, car, bus}\} \\
\midrule
Personal Transport & \texttt{person} $\wedge$ (Either \texttt{bicycle} or \texttt{car}) \\
\midrule
Surf Trip & Exactly as many \texttt{person} as \texttt{surfboard} $\wedge$ at least one of each\\
\bottomrule
\end{tabular}%
}
\end{table}

Existing evaluations of interpretable deep learning models fall short of real-world complexity: CBMs are mainly assessed on single-label classification benchmarks \cite{welinder2010caltech,deng2009imagenet,zhou2017places}, while program synthesis and neuro-symbolic approaches rely on synthetic tasks \cite{muller2021kandinsky,vedantam2021curi,StammerSK21} or simple concept identification rather than complex reasoning \cite{wu2024bongard,jiang2022bongard}. 
We introduce \data as a step towards more complex tasks and an evaluation for inductive reasoning over real-world images.
\data builds on COCOLogic \cite{steinmann2026object}, expanding the logical scope to include object counting and counting comparisons and reformulating the task as multilabel classification to reduce class imbalance and shortcuts. Most importantly, a new sampling procedure introduces positive variants and near-boundary negatives for automatic model diagnosis. We also introduce \dataf, a curated version for few-shot and in-context rule learning on complex real-world images.

\textbf{Compositional Logical Rules.}
\data includes 10 logical rules based on object co-occurrences in MSCOCO \cite{lin2014microsoft} images. The rules are semantically meaningful and cover a subset of first-order logic, including propositional logic, object counting, counting comparisons, and object absence. The full rule list is shown in \autoref{tab:rules}.
Beyond binary labels, \data provides fine-grained annotations for model diagnosis: positive samples are split into \textit{positive variants} and negatives into \textit{near-boundary (NB)} types and \textit{far-from-boundary (FB)} samples. 
The positive variants describe the different ways a rule can be satisfied, while the NB types represent samples near the decision boundary, providing specific information about how well a model has learned the rule's details.

\begin{figure}[t]
    \centering
    \resizebox{0.9\linewidth}{!}{\begin{tikzpicture}[
  font=\small,
  rulebox/.style={
    draw=black!75, ultra thick, rounded corners=5pt,
    fill=white, minimum width=14cm, align=center, inner sep=5pt
  },
  dbox/.style={
    draw=#1, ultra thick, rounded corners=3pt, fill=none,
    align=center, inner sep=5pt
  },
  pvbox/.style={
    draw=#1, ultra thick, rounded corners=3pt, fill=none,
    align=center, inner sep=5pt
  },
  nbbox/.style={
    draw=#1, ultra thick, rounded corners=3pt, fill=none,
    align=center, inner sep=5pt,
    minimum width=4.3cm
  },
  sechdr/.style={font=\small\bfseries, align=center},
  secbg/.style={
    draw=gray!55, dashed, rounded corners=5pt,
    fill=gray!5, inner sep=5pt
  },
  main_arr/.style={->, >=Stealth, thick},
  col_arr/.style={->, >=Stealth, semithick, draw=#1!70!black},
  dash_arr/.style={->, >=Stealth, semithick, dashed, draw=gray!65},
]

\node[rulebox] (rule) at (0, 0) {%
  Exactly two categories of
  $\bigl\{\,\mathtt{bottle},\;\mathtt{cup},\;\mathtt{pizza}\,\bigr\}$ must be present.%
};

\node[dbox=ColI,   below=0.8cm of rule, xshift=-4.8cm] (d1)
  {$\mathtt{bottle} \wedge \mathtt{cup} \wedge \neg\,\mathtt{pizza}$};
\node[dbox=ColII,  below=0.8cm of rule                ] (d2)
  {$\mathtt{bottle} \wedge \neg\,\mathtt{cup} \wedge \mathtt{pizza}$};
\node[dbox=ColIII, below=0.8cm of rule, xshift= 4.8cm] (d3)
  {$\neg\,\mathtt{bottle} \wedge \mathtt{cup} \wedge \mathtt{pizza}$};

\node[font=\large] at ($(d1.east)!.5!(d2.west)$) {$\vee$};
\node[font=\large] at ($(d2.east)!.5!(d3.west)$) {$\vee$};

\begin{scope}[on background layer]
  \node[secbg, inner xsep=10pt, inner ysep=5pt, fit=(d1)(d2)(d3)]
    (dnfbg) {};
\end{scope}
\node[below=0.08cm of rule, xshift=0.55cm, font=\small, text=black, align=right]
  {\footnotesize DNF};

\draw[main_arr] (rule.south) -- (dnfbg.north);

\node[sechdr, below=0.8cm of dnfbg, xshift=-4.85cm] (pvhdr)
  {Positive Variants};

\node[pvbox=ColI,   below=0.15cm of pvhdr] (pv1)
  {$\mathtt{bottle} \wedge \mathtt{cup} \wedge \neg\,\mathtt{pizza}$};
\node[pvbox=ColII,  below=0.2cm of pv1  ] (pv2)
  {$\mathtt{bottle} \wedge \neg\,\mathtt{cup} \wedge \mathtt{pizza}$};
\node[pvbox=ColIII, below=0.2cm of pv2  ] (pv3)
  {$\neg\,\mathtt{bottle} \wedge \mathtt{cup} \wedge \mathtt{pizza}$};

\begin{scope}[on background layer]
  \node[secbg, fill=blue!3, fit=(pvhdr)(pv1)(pv2)(pv3)] (pvbg) {};
\end{scope}

\draw[main_arr] ([shift={(-4.85cm,0.01cm)}]dnfbg.south) -- ([shift={(0cm,+0.18cm)}]pvhdr.north);
\node[below=0.1cm of dnfbg, xshift=-3.95cm, font=\small, text=black, align=right]
  {\footnotesize Enumerate};

\node[sechdr, below=0.8cm of dnfbg, xshift=2.35cm] (nbhdr)
  {Near-Boundary (NB) Types};

\node[nbbox=gray!80] (nb1) at ($(pv1)+(4.85cm,0)$)
  {$\mathtt{bottle} \;\wedge\;
    \textcolor{ColFlip}{\boldsymbol{\neg\,\mathtt{cup}}} \;\wedge\;
    \neg\,\mathtt{pizza}$};

\node[nbbox=gray!80, right=0.2cm of nb1] (nb2)
  {$\textcolor{ColFlip}{\boldsymbol{\neg\,\mathtt{bottle}}} \;\wedge\;
    \neg\,\mathtt{cup} \;\wedge\; \mathtt{pizza}$};

\node[nbbox=gray!80] (nb3) at (nb1 |- pv2)
  {$\neg\,\mathtt{bottle} \;\wedge\;
    \mathtt{cup} \;\wedge\; \textcolor{ColFlip}{\boldsymbol{\neg\,\mathtt{pizza}}}$};

\node[nbbox=gray!80] (nb4) at (nb2 |- pv2)
  {$\mathtt{bottle} \;\wedge\; \mathtt{cup} \;\wedge\;
    \textcolor{ColFlip}{\boldsymbol{\mathtt{pizza}}}$};

\begin{scope}[on background layer]
  \node[secbg, fill=orange!8, fit=(nbhdr)(nb1)(nb2)(nb3)(nb4)] (nbbg) {};
\end{scope}

\draw[main_arr] ([shift={(2.35cm,0.01cm)}]dnfbg.south) -- ([shift={(0cm,+0.18cm)}]nbhdr.north);
\node[below=0.1cm of dnfbg, xshift=+3.1cm, font=\small, text=black, align=right]
  {\footnotesize Permute};

\end{tikzpicture}}
    \caption{\textbf{Obtaining Positive Variants and Near-Boundary (NB) Types.} First, the logic rule (top) is brought to its disjunctive normal form (DNF). Then each disjunct represents one positive variant. The NB types are obtained by permuting the disjuncts of the DNF, for example, by negating a literal.}
    \label{fig:variant_construction}
    \vspace{-0.5cm}
\end{figure}
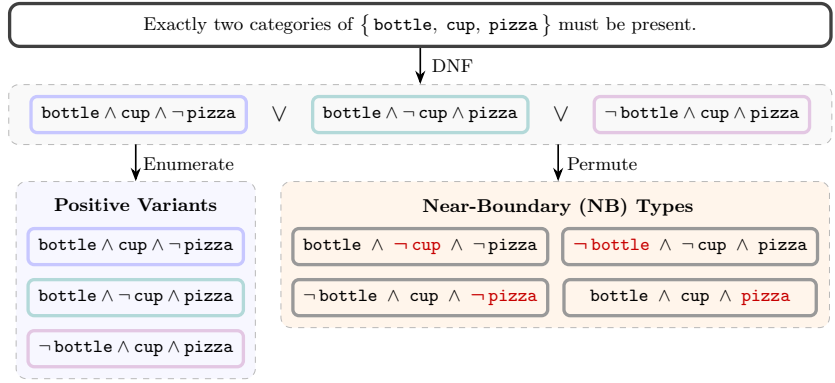

The process to obtain positive variants and NB types from a rule is illustrated in \autoref{fig:variant_construction}. First, it is brought to its disjunctive normal form (DNF).
Each positive variant corresponds to one disjunct of the DNF, so that the full positive set is the union of all variants.
For "Double Serving", exactly two categories of \{\texttt{bottle}, \texttt{cup}, \texttt{pizza}\} must be present, yielding the positive variants:
\mbox{$(\texttt{bottle} \wedge \texttt{cup} \wedge \neg\texttt{pizza})$}, \mbox{$(\texttt{bottle} \wedge \neg\texttt{cup}\wedge \texttt{pizza})$}, and \mbox{$(\neg\texttt{bottle} \wedge \texttt{cup} \wedge \texttt{pizza})$}.
Each NB type is derived from a DNF disjunct by flipping one literal, or by modifying object counts for counting rules. For "Double Serving," this yields four different NB types: three single-category variants \mbox{$(\texttt{bottle} \wedge \neg \texttt{cup} \wedge \neg \texttt{pizza})$},
\mbox{$(\neg \texttt{bottle} \wedge \neg \texttt{cup} \wedge \texttt{pizza})$}, 
\mbox{$(\neg \texttt{bottle} \wedge \texttt{cup} \wedge \neg \texttt{pizza})$}, 
and the all-three variant \mbox{$(\texttt{bottle} \wedge \texttt{cup} \wedge \texttt{pizza})$}.
We list every NB type and positive variant that can be derived in \autoref{app:rule_details}.

\textbf{Sampling \data from MSCOCO.} Since \data is based on images from MSCOCO, and most of these images are negative for all rules, we sample 25,000 images to achieve a more balanced distribution.
Sampling proceeds in three steps, applied identically to training and test sets (with test images drawn from MSCOCO's validation split).
First, up to 1,000 images per positive variant are randomly selected, with cross-rule duplicates removed where this does not drop any variant below 1,000 samples. 
Second, images for each NB type are increased to at least 500 (if enough samples are available). 
Third, FB samples drawn from the remaining MSCOCO images are added to keep \data's overall distribution close to MSCOCO's. The resulting training set contains 25,000 images and the test set 3,500 (see \autoref{app:sampling_details} for details). All images are annotated with rule labels, positive variant identifiers, and NB type or FB status for negative samples.

\textbf{\dataf.} This curated subset of \data is designed for few-shot and in-context learning. 
With only 24 training samples per rule, efficient rule learning is essential to succeed on this version. 
\dataf assumes a functioning perception module and is not designed to train such a module from scratch. 
All images were manually selected to ensure relevant objects are clearly visible and MSCOCO labels are correct.
\dataf contains 239 training and 368 test images in total. 
Per rule, the training set contains 8 positive and 16 negative samples and the test set contains 20 of each. 
Positive images are evenly distributed across variants, and negative images include two fixed FB samples with the remainder spread evenly across NB types (cf. \autoref{app:sampling_details} for detailed sample distributions).

\section{Experimental Evaluations}\label{sec:eval}
In this section, we evaluate several interpretable approaches alongside black-box baselines to assess the current state of visual inductive reasoning on \data. We aim to answer two questions: (1) How well do current concept-based approaches perform on visual inductive reasoning? (2) What are the key bottlenecks for visual inductive reasoning in the few-shot setting?

\textbf{Models.} On the full dataset, we evaluate a fine-tuned ResNet-50 \cite{he2016deep} as a black-box baseline, and a range of CBM variants with linear or MLP predictors: using ground-truth object labels (Oracle), a pretrained Mask-RCNN \cite{he2017mask} detector, a supervised ResNet-50 concept encoder \cite{koh2020concept}, CLIP-based concept representations \cite{yang2023language}, object-centric CBMs (OCB) \cite{steinmann2026object}, and Deep Concept Reasoner (DCR) \cite{barbiero2023interpretable}. 
On \dataf, we additionally evaluate GPT-5.5 \cite{gpt5_5}, Gemini-3.1-Pro-Preview \cite{gemini3_1_pro}, and Claude-Opus-4-7 \cite{claude_opus_4_7} for in-context learning, and the VLP program synthesis approach \cite{wust2026synthesizing} with Gemma-4-31B-it \cite{gemma_4} as backbone. Further implementation details are in \autoref{app:experimental_details}.

\textbf{Metrics.} All metrics are computed per rule and averaged across rules. We report balanced accuracy (B-Acc) as an overall summary, alongside separate accuracies on positive variants and NB types. 
The latter two are more informative than raw B-Acc, as they are not inflated by easy FB samples and show directly whether a model has learned the logical rules.

\begin{table}[t]
\centering
\caption{\textbf{Concept-based Models fail to learn details of logical rules on \data.} High overall B-Acc paired with low performance on NB types shows that models learn a broad separation between positive and easy FB samples, but fail when challenged with samples that require a proper understanding of the logic rule. 
Best bold, runner up underlined, mean $\pm$ std over 5 seeds.}
\label{tab:cocologicv2-full}
\resizebox{0.8\linewidth}{!}{%
\begin{tabular}{l l r r r}
\toprule
Encoder & Predictor & B-Acc & NB Types & Positive Variants \\
\midrule
Oracle & Linear & $94.0 \pm 0.1$ & $44.1 \pm 0.3$ & $96.0 \pm 0.1$ \\
Oracle & MLP & \textbf{$99.6 \pm 0.1$} & \textbf{$94.3 \pm 2.5$} & \textbf{$99.6 \pm 0.2$} \\
\midrule
ResNet-50 & -- & $86.8 \pm 0.4$ & $40.5 \pm 1.4$ & $86.5 \pm 1.2$ \\
\midrule
MaskRCNN & Linear & $\underline{88.8 \pm 0.1}$ & $42.0 \pm 0.4$ & $89.1 \pm 0.3$ \\
MaskRCNN & MLP & $\mathbf{92.1 \pm 0.2}$ & $\mathbf{48.5 \pm 0.8}$ & $\mathbf{92.9 \pm 0.3}$ \\
\addlinespace[0.2em]
Supervised & Linear & $86.0 \pm 0.1$ & $30.8 \pm 0.7$ & $89.6 \pm 0.3$ \\
Supervised & MLP & $86.7 \pm 0.1$ & $32.9 \pm 0.5$ & $90.9 \pm 0.6$ \\
\addlinespace[0.2em]
DCR & DCR & $82.3 \pm 3.6$ & $\mathbf{48.5 \pm 8.6}$ & $76.6 \pm 9.1$ \\
\addlinespace[0.2em]
CLIP & Linear & $84.7 \pm 0.1$ & $28.1 \pm 0.2$ & $87.8 \pm 0.3$ \\
CLIP & MLP & $86.5 \pm 0.2$ & $27.1 \pm 1.0$ & $\underline{92.2 \pm 0.7}$ \\
\addlinespace[0.2em]
OCB & Linear & $85.8 \pm 0.1$ & $31.2 \pm 0.4$ & $88.9 \pm 0.5$ \\
OCB & MLP & $86.2 \pm 0.3$ & $31.3 \pm 1.2$ & $90.4 \pm 0.5$ \\
\bottomrule
\end{tabular}
}
\vspace{-0.5cm}
\end{table}

\textbf{Concept-Based Approaches on \data (RQ1).} The results in \autoref{tab:cocologicv2-full} appear promising at first: all models exceed 80\% B-Acc overall. However, NB accuracy reveals a fundamental shortcoming, as all models except Oracle + MLP perform at or below random guessing on NB samples. Rather than learning the logical rules, models separate positives from easy FB samples based on the coarse presence of relevant objects. 
NB accuracy degrades further as concept representations become noisier: the pretrained detector outperforms the supervised encoder, and CLIP-based representations perform worst. This suggests that fine-grained rule details become increasingly difficult to capture with lower-quality concepts. 
While DCR's logic-based predictor is, in principle, well-suited for these tasks, it is hindered by noisy perception and a large concept space.
The Oracle upper bound confirms the dataset is solvable in principle: a linear predictor alone fails on rules involving XOR or exact counting, but an MLP achieves near-perfect performance.
The detailed results per rule further show that overall and NB accuracy are largely independent across rules (cf. \autoref{app:result_details}).

\begin{table}[t]
\centering
\caption{\textbf{Combining Perception and Rule Induction on few, real-world images remains challenging.} Training CBM Predictors on \dataf per-rule gives low performance and unstable results. In contrast, in-context learning and program synthesis perform better, but identifying what objects are present and relevant to find the logical rules remains a key challenge.}
\label{tab:cocologicv2-per-rule}
\resizebox{0.8\linewidth}{!}{%
\begin{tabular}{l rrr}
\toprule
Model & B-Acc & NB Types & Positive Variants \\
\midrule
Oracle + Linear & $58.0 \pm 2.4$ & $56.6 \pm \phantom{0}8.1$ & $55.7 \pm \phantom{0}8.4$ \\
Oracle + MLP & \textbf{$62.8 \pm 1.9$} & \textbf{$62.9 \pm 13.4$} & \textbf{$59.0 \pm 13.2$} \\
\midrule
ResNet-50  & $62.3 \pm 1.6$ & $74.9 \pm \phantom{0}4.2$ & $47.0 \pm \phantom{0}4.4$ \\
\midrule
MaskRCNN + Linear & $51.3 \pm 1.9$ & $68.0 \pm \phantom{0}3.6$ & $35.9 \pm \phantom{0}5.6$ \\
MaskRCNN + MLP & $49.5 \pm 2.1$ & $60.2 \pm 27.7$ & $38.6 \pm 23.9$ \\
\addlinespace[0.2em]
Supervised + Linear & $60.8 \pm 2.6$ & $50.6 \pm \phantom{0}6.1$ & $\mathbf{67.6 \pm \phantom{0}6.7}$ \\
Supervised + MLP & $\mathbf{61.7 \pm 2.3}$ & $57.0 \pm 19.3$ & $62.8 \pm 21.5$ \\
\addlinespace[0.2em]
DCR & $52.3 \pm 3.1$ & $\underline{69.1 \pm 16.2}$ & $36.5 \pm 19.6$ \\
\addlinespace[0.2em]
CLIP + Linear & $55.8 \pm 2.9$ & $47.7 \pm 12.9$ & $58.6 \pm 16.0$ \\
CLIP + MLP & $\underline{61.5 \pm 1.4}$ & $53.2 \pm \phantom{0}1.3$ & $\underline{62.9 \pm \phantom{0}2.4}$ \\
\addlinespace[0.2em]
OCB + Linear & $52.4 \pm 1.3$ & $64.7 \pm \phantom{0}2.5$ & $40.6 \pm \phantom{0}3.0$ \\
OCB + MLP & $51.1 \pm 1.7$ & $\mathbf{69.8 \pm \phantom{0}5.2}$ & $31.6 \pm \phantom{0}5.7$ \\
\midrule
\midrule
\multicolumn{4}{l}{\textit{In-Context Learning \& Program Synthesis}} \\
VLP (Oracle) & $89.2 \pm 0.0$ & $90.6 \pm \phantom{0}0.0$ & $88.4 \pm \phantom{0}0.0$ \\
\midrule
VLP (Gemma-4-31B-it) & $72.3 \pm 1.3$ & $\underline{74.5 \pm \phantom{0}2.1}$ & $\underline{66.1 \pm \phantom{0}3.5}$ \\
Gemma-4-31B-it & $65.4 \pm 2.3$ & $60.7 \pm \phantom{0}4.7$ & $65.6 \pm \phantom{0}6.8$ \\ 
GPT-5.5 & $68.1 \pm 1.4$ & $74.3 \pm \phantom{0}3.5$ & $56.3 \pm \phantom{0}3.8$ \\
Gemini-3.1-Pro-Preview & $\underline{74.8 \pm 2.9}$ & $62.7 \pm \phantom{0}8.1$ & $\mathbf{78.3 \pm \phantom{0}3.4}$ \\
Claude-Opus-4-7 & $\mathbf{75.5 \pm 2.4}$ & $\mathbf{80.5 \pm \phantom{0}4.8}$ & $65.9 \pm \phantom{0}5.2$ \\
\bottomrule
\end{tabular}
}
\vspace{-0.5cm}
\end{table}

\textbf{Visual Inductive Reasoning in Few-Shot Settings (RQ2).} We now examine the few-shot setting to identify where current approaches break down when training signal is scarce. We observe in \autoref{tab:cocologicv2-per-rule} that all CBM approaches struggle on \dataf: even Oracle + MLP reaches only 62.8\% B-Acc with high variance across seeds. 
The limited training signal from 24 samples per rule makes it difficult to learn reliable predictors, and differences between encoder types are largely overshadowed by high variance. 
Interestingly, NB accuracy tends to be higher here than on the full dataset, as models do not systematically misclassify NB samples without sufficient data to overfit to positive vs. FB differences.
In contrast, in-context learning with VLMs offers a clear advantage over CBMs, with all three tested models reaching around 70\% B-Acc. 
In some cases, they recover the correct rule, as Claude produces the exact ground-truth description for Signal and Ride. However, for rules where identifying relevant objects is non-trivial, VLMs fall short: on Double Serving, Claude describes a semantically related but logically imprecise pattern rather than the exact counting constraint. 
Program synthesis with VLP (Gemma-4-31B-it) performs comparably to the best VLMs and benefits from an explicit structure for rule induction, despite the lower baseline performance of Gemma compared to other VLMs.
Providing ground-truth concepts (VLP Oracle) boosts performance substantially to 88.5\%, but even then, complex rules such as Single Mode Traffic remain difficult to identify within constrained search time. 
Together, these results suggest that reliable identification of which objects are present and decisions about which are relevant remain the primary bottleneck for few-shot visual inductive reasoning.

\section{Discussion}\label{sec:discussion}

Our evaluations on \data show that visual logical reasoning on real-world images remains a significant open challenge. On the full dataset, models achieve high overall accuracy by separating positives from easy FB samples, but fail on NB samples, indicating they exploit coarse statistical patterns rather than learning the underlying rules. The annotated sample categories enable automatic diagnosis of such shortcomings without detailed human inspection: a model that fails on NB samples while achieving high overall accuracy is clearly not learning the intended rule. In few-shot settings, perceptual noise and large rule-induced search spaces pose additional challenges that even specialized approaches such as program synthesis do not fully overcome. Together, these results validate the need for a benchmark like \data that goes beyond aggregate accuracy to reveal where and why current interpretable models fall short.

However, NB annotations do not entirely replace human inspection. Models can still exploit other shortcuts not captured by the NB structure, such as using the presence of water to predict the Surf Trip rule. Mitigating general \cite{steinmann2024navigating} and reasoning-specific shortcuts \cite{marconato2023not} thus remains important to ensure accountability.

\section{Conclusion}\label{sec:conclusion}

We introduced \data, a dataset for visual inductive reasoning on real-world images, along with a curated few-shot version \dataf. Our evaluations show that current approaches, including CBMs, VLMs, and program synthesis, struggle to capture the underlying logical rules, particularly on near-boundary samples, highlighting that visual logical reasoning on real-world data remains an open challenge. 
\data provides a useful foundation for advancing and benchmarking methods in this direction.
Going forward, extending the dataset to include more complex logical structures, object relations, and more abstract scene-level information is a promising direction that will further test the limits of interpretable models.

\begin{credits}
\subsubsection{\ackname}
This work was supported by the ”ML2MT” project from the Volkswagen Stiftung, by the German Research Foundation (DFG) under Germany’s Excellence Strategy (EXC 3066/1 “The Adaptive Mind”, Project No. 533717223; and GRK 2853 "Neuroexplicit Models of Language, Vision, and Action", Project No. 471607914). It was further supported by the EU-funded “TANGO” project (EU Horizon 2023, GA No 57100431), and the Priority Program (SPP) 2422 in the subproject “Optimization of active surface design of high-speed progressive tools using machine and deep learning algorithms“ funded by the German Research Foundation (DFG).
It has benefited from the HMWK project Hessian.AI, and from the Cluster of Excellence "Reasonable AI" funded by the DFG under Germany’s Excellence Strategy EXC-3057.

\subsubsection{\discintname}
The authors have no competing interests to declare that are
relevant to the content of this article.
\end{credits}

\bibliographystyle{splncs04}
\bibliography{sample}

@inproceedings{koh2020concept,
  title={Concept bottleneck models},
  author={Koh, Pang Wei and Nguyen, Thao and Tang, Yew Siang and Mussmann, Stephen and Pierson, Emma and Kim, Been and Liang, Percy},
  booktitle={International Conference on Machine Learning (ICML)},
  pages={5338--5348},
  year={2020},
}

@inproceedings{StammerSK21,
  author       = {Wolfgang Stammer and
                  Patrick Schramowski and
                  Kristian Kersting},
  title        = {Right for the Right Concept: Revising Neuro-Symbolic Concepts by Interacting
                  With Their Explanations},
  booktitle    = {Conference on Computer Vision and Pattern Recognition {CVPR}},
  pages        = {3619--3629},
  year         = {2021},
}

@article{espinosa2022concept,
  title={Concept embedding models: Beyond the accuracy-explainability trade-off},
  author={Espinosa Zarlenga, Mateo and Barbiero, Pietro and Ciravegna, Gabriele and Marra, Giuseppe and Giannini, Francesco and Diligenti, Michelangelo and Shams, Zohreh and Precioso, Frederic and Melacci, Stefano and Weller, Adrian and others},
  journal={Advances in neural information processing systems},
  volume={35},
  pages={21400--21413},
  year={2022}
}

@article{ellis2023dreamcoder,
  title={Dreamcoder: growing generalizable, interpretable knowledge with wake--sleep bayesian program learning},
  author={Ellis, Kevin and Wong, Lionel and Nye, Maxwell and Sable-Meyer, Mathias and Cary, Luc and Anaya Pozo, Lore and Hewitt, Luke and Solar-Lezama, Armando and Tenenbaum, Joshua B},
  journal={Philosophical Transactions of the Royal Society A: Mathematical, Physical and Engineering Sciences},
  volume={381},
  number={2251},
  year={2023},
  publisher={The Royal Society}
}

@inproceedings{wust2026synthesizing,
  title={Synthesizing visual concepts as vision-language programs},
  author={W{\"u}st, Antonia and Stammer, Wolfgang and Shindo, Hikaru and Helff, Lukas and Dhami, Devendra Singh and Kersting, Kristian},
  booktitle={Proceedings of the IEEE/CVF Conference on Computer Vision and Pattern Recognition},
  pages={17346--17356},
  year={2026}
}

@article{knab2026bottle,
  title={What’s in the Bottle? A Survey and Roadmap of Concept Bottleneck Models},
  author={Knab, Patrick and Steinmann, David and Bartelt, Christian and Kersting, Kristian and Schiele, Bernt and Seidl, Thomas and Schlegel, Udo and Stammer, Wolfgang},
  journal={Transactions on Machine Learning Research},
year={2026},
}

@article{steinmann2026object,
  title={Object-centric concept-bottlenecks},
  author={Steinmann, David and Stammer, Wolfgang and W{\"u}st, Antonia and Kersting, Kristian},
  journal={Advances in Neural Information Processing Systems},
  volume={38},
  pages={68899--68924},
  year={2026}
}

@inproceedings{LinMBHPRDZ14,
  author       = {Tsung{-}Yi Lin and
                  Michael Maire and
                  Serge J. Belongie and
                  James Hays and
                  Pietro Perona and
                  Deva Ramanan and
                  Piotr Doll{\'{a}}r and
                  C. Lawrence Zitnick},
  title        = {Microsoft {COCO:} Common Objects in Context},
  booktitle    = {European Conference on Computer Vision (ECCV)},
  pages        = {740--755},
  year         = {2014},
}

@article{Rudin19,
  author       = {Cynthia Rudin},
  title        = {Stop explaining black box machine learning models for high stakes
                  decisions and use interpretable models instead},
  journal      = {Nature Machine Intelligence},
  volume       = {1},
  number       = {5},
  pages        = {206--215},
  year         = {2019},
}

@article{welinder2010caltech,
  title={Caltech-UCSD birds 200},
  author={Welinder, Peter and Branson, Steve and Mita, Takeshi and Wah, Catherine and Schroff, Florian and Belongie, Serge and Perona, Pietro},
  year={2010}
}

@inproceedings{deng2009imagenet,
  title={Imagenet: A large-scale hierarchical image database},
  author={Deng, Jia and Dong, Wei and Socher, Richard and Li, Li-Jia and Li, Kai and Fei-Fei, Li},
  booktitle={2009 IEEE conference on computer vision and pattern recognition},
  pages={248--255},
  year={2009},
  organization={Ieee}
}

@article{zhou2017places,
  title={Places: A 10 million image database for scene recognition},
  author={Zhou, Bolei and Lapedriza, Agata and Khosla, Aditya and Oliva, Aude and Torralba, Antonio},
  journal={IEEE transactions on pattern analysis and machine intelligence},
  volume={40},
  number={6},
  pages={1452--1464},
  year={2017},
  publisher={IEEE}
}

@inproceedings{jiang2022bongard,
  title={Bongard-hoi: Benchmarking few-shot visual reasoning for human-object interactions},
  author={Jiang, Huaizu and Ma, Xiaojian and Nie, Weili and Yu, Zhiding and Zhu, Yuke and Anandkumar, Anima},
  booktitle={Proceedings of the IEEE/CVF conference on computer vision and pattern recognition},
  pages={19056--19065},
  year={2022}
}

@inproceedings{wu2024bongard,
  title={Bongard-openworld: Few-shot reasoning for free-form visual concepts in the real world},
  author={Wu, Rujie and Ma, Xiaojian and Zhang, Zhenliang and Wang, Wei and Li, Qing and Zhu, Song-Chun and Wang, Yizhou},
  booktitle={International Conference on Learning Representations},
  volume={2024},
  pages={20688--20718},
  year={2024}
}

@inproceedings{vedantam2021curi,
  title={Curi: A benchmark for productive concept learning under uncertainty},
  author={Vedantam, Ramakrishna and Szlam, Arthur and Nickel, Maximillian and Morcos, Ari and Lake, Brenden M},
  booktitle={International Conference on Machine Learning},
  pages={10519--10529},
  year={2021},
  organization={PMLR}
}

@inproceedings{lin2014microsoft,
  title={Microsoft coco: Common objects in context},
  author={Lin, Tsung-Yi and Maire, Michael and Belongie, Serge and Hays, James and Perona, Pietro and Ramanan, Deva and Doll{\'a}r, Piotr and Zitnick, C Lawrence},
  booktitle={European conference on computer vision},
  pages={740--755},
  year={2014},
  organization={Springer}
}

@inproceedings{he2016deep,
  title={Deep residual learning for image recognition},
  author={He, Kaiming and Zhang, Xiangyu and Ren, Shaoqing and Sun, Jian},
  booktitle={Proceedings of the IEEE conference on computer vision and pattern recognition},
  pages={770--778},
  year={2016}
}

@inproceedings{he2017mask,
  title={Mask r-cnn},
  author={He, Kaiming and Gkioxari, Georgia and Doll{\'a}r, Piotr and Girshick, Ross},
  booktitle={Proceedings of the IEEE international conference on computer vision},
  pages={2961--2969},
  year={2017}
}

@inproceedings{yang2023language,
  title={Language in a bottle: Language model guided concept bottlenecks for interpretable image classification},
  author={Yang, Yue and Panagopoulou, Artemis and Zhou, Shenghao and Jin, Daniel and Callison-Burch, Chris and Yatskar, Mark},
  booktitle={Proceedings of the IEEE/CVF conference on computer vision and pattern recognition},
  pages={19187--19197},
  year={2023}
}

@inproceedings{barbiero2023interpretable,
  title={Interpretable neural-symbolic concept reasoning},
  author={Barbiero, Pietro and Ciravegna, Gabriele and Giannini, Francesco and Zarlenga, Mateo Espinosa and Magister, Lucie Charlotte and Tonda, Alberto and Li{\'o}, Pietro and Precioso, Frederic and Jamnik, Mateja and Marra, Giuseppe},
  booktitle={International Conference on Machine Learning},
  pages={1801--1825},
  year={2023},
  organization={PMLR}
}

@article{steinmann2024navigating,
  title={Navigating shortcuts, spurious correlations, and confounders: From origins via detection to mitigation},
  author={Steinmann, David and Divo, Felix and Kraus, Maurice and W{\"u}st, Antonia and Struppek, Lukas and Friedrich, Felix and Kersting, Kristian},
  journal={arXiv preprint arXiv:2412.05152},
  year={2024}
}

@article{marconato2023not,
  title={Not all neuro-symbolic concepts are created equal: Analysis and mitigation of reasoning shortcuts},
  author={Marconato, Emanuele and Teso, Stefano and Vergari, Antonio and Passerini, Andrea},
  journal={Advances in Neural Information Processing Systems},
  volume={36},
  pages={72507--72539},
  year={2023}
}

@inproceedings{mao2019neuro,
  title={The Neuro-Symbolic Concept Learner: Interpreting Scenes, Words, and Sentences From Natural Supervision},
  author={Mao, Jiayuan and Gan, Chuang and Kohli, Pushmeet and Tenenbaum, Joshua B and Wu, Jiajun},
  booktitle={International Conference on Learning Representations},
    year={2019}
}

@inproceedings{
kamali2025neptune,
title={Ne{PT}une: A Neuro-Pythonic Framework for Tunable Compositional Reasoning on Vision-Language},
author={Danial Kamali and Parisa Kordjamshidi},
booktitle={The Fourteenth International Conference on Learning Representations},
year={2026},
}

@inproceedings{VisProg,
  author    = {Gupta, Tanmay and Kembhavi, Aniruddha},
  title     = {Visual Programming: Compositional visual reasoning without training},
  booktitle = {{IEEE/CVF} Conference on Computer Vision and Pattern Recognition},
  year      = {2023}
}

@article{chollet2019measure,
  title={On the measure of intelligence},
  author={Chollet, Fran{\c{c}}ois},
  journal={arXiv preprint arXiv:1911.01547},
  year={2019}
}

@misc{gemini3_1_pro,
  title        = {Gemini 3.1 Pro (Preview)},
  author       = {{Google DeepMind}},
  year         = {2026},
  howpublished = {Model card},
  url          = {https://deepmind.google/models/gemini/pro/}
}

@misc{gpt5_5,
  title        = {GPT-5.5},
  author       = {{OpenAI}},
  year         = {2026},
  howpublished = {System card},
  url          = {https://openai.com/index/gpt-5-5-system-card/}
}

@misc{claude_opus_4_7,
  title        = {Claude Opus 4.7},
  author       = {{Anthropic}},
  year         = {2026},
  howpublished = {Model card},
  url          = {https://www.anthropic.com/claude/opus}
}

@misc{gemma_4,
  title        = {Gemma 4},
  author       = {{Google DeepMind}},
  year         = {2026},
  howpublished = {Model card},
  url          = {https://deepmind.google/models/gemma/gemma-4/}
}

@article{shindo2023alpha,
  title={$\alpha$ ilp: thinking visual scenes as differentiable logic programs},
  author={Shindo, Hikaru and Pfanschilling, Viktor and Dhami, Devendra Singh and Kersting, Kristian},
  journal={Machine Learning},
  volume={112},
  number={5},
  pages={1465--1497},
  year={2023},
  publisher={Springer}
}

@article{panousis2024coarse,
  title={Coarse-to-fine concept bottleneck models},
  author={Panousis, Konstantinos P and Ienco, Dino and Marcos, Diego},
  journal={Advances in Neural Information Processing Systems},
  volume={37},
  pages={105171--105199},
  year={2024}
}

@inproceedings{shang2024incremental,
  title={Incremental residual concept bottleneck models},
  author={Shang, Chenming and Zhou, Shiji and Zhang, Hengyuan and Ni, Xinzhe and Yang, Yujiu and Wang, Yuwang},
  booktitle={Proceedings of the IEEE/CVF Conference on Computer Vision and Pattern Recognition},
  pages={11030--11040},
  year={2024}
}

@article{muller2021kandinsky,
  title={Kandinsky patterns},
  author={M{\"u}ller, Heimo and Holzinger, Andreas},
  journal={Artificial intelligence},
  volume={300},
  pages={103546},
  year={2021},
  publisher={Elsevier}
}

\onecolumn
\begin{center}
\textbf{\large Supplementary Materials}
\end{center}
\setcounter{section}{0}
\renewcommand{\thesection}{\Alph{section}}

\section{Additional Details for \data}
\subsection{Positive Variants and Near Boundary Types}
\label{app:rule_details}
\begin{figure}
    \centering
    \includegraphics[width=\linewidth]{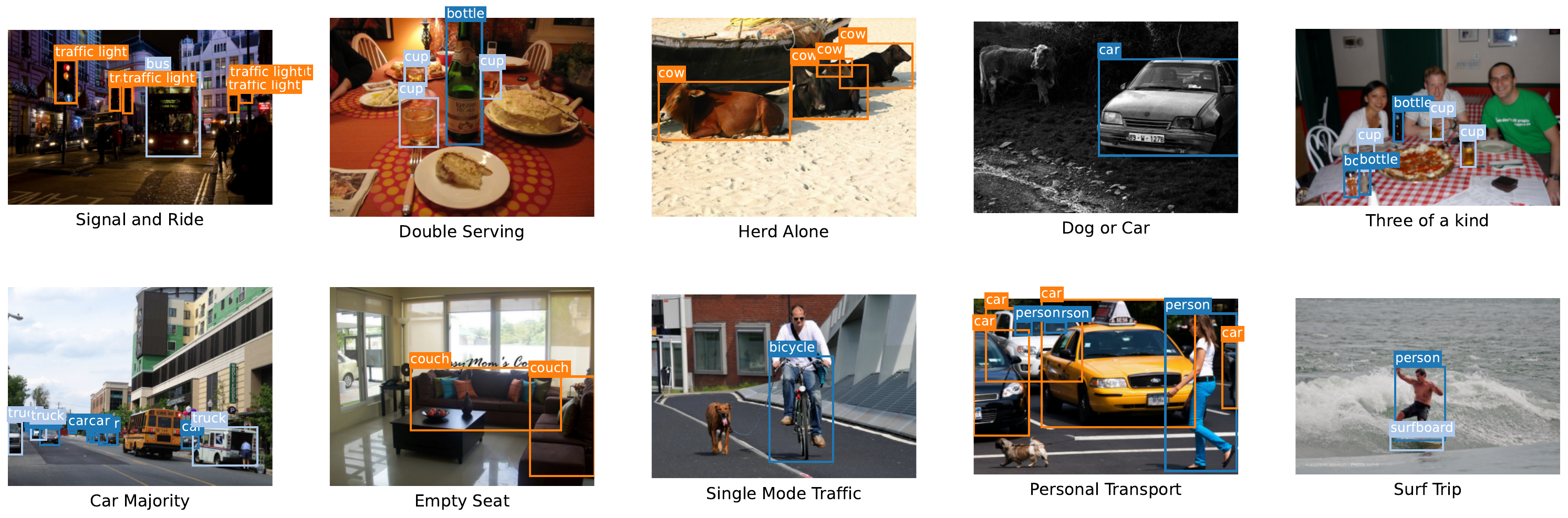}
    \caption{Example images for all rules of \data with the relevant objects highlighted.}
    \label{fig:examples_all_rules}
\end{figure}

\data consists of 10 rules, each with several positive variants and near-boundary (NB) types. In \autoref{fig:examples_all_rules}, an example image of each rule is given, where the objects that are relevant to the respective rule are highlighted. These examples highlight the wide variety and difficulty of identifying potential objects and deciding which might be relevant to the task, compared to synthetic datasets that are often clearer and contain less potentially relevant information.

In the following, we provide a detailed list of all positive variants and NB types for all rules. The positive variants are enumerated with \textbf{P}$i$ and the NB types are enumerated with \textbf{N}$i$. In cases where rules require object counting, we simplify the logical structure for object counts and assume the presence of a predicate that provides the number of times an object appears in an image, for further use in the logical formulation. Following this, the NB types for these rules modify the required object counts instead of directly flipping a literal (e.g., for rule 5, where \textbf{N1} is "1 -- 2 \texttt{cup}, rather than "$\neg$ exactly 3 \texttt{cup}").

\newcommand{\pitem}[2]{\textbf{P#1} & #2}
\newcommand{\nitem}[2]{\textbf{N#1} & #2}

\tcbset{
  rulebox/.style={
    enhanced, breakable,
    colback=white, colframe=black!40,
    fonttitle=\footnotesize,
    fontupper=\scriptsize, 
    attach boxed title to top left={yshift=-3mm, xshift=4mm},
    boxed title style={colback=black!70, colframe=black!70, fontupper=\color{white}},
    top=4mm, left=2mm, right=2mm, bottom=2mm,
    before skip=8pt, after skip=8pt,
  }
}

\setlength{\tabcolsep}{5pt}
\renewcommand{\arraystretch}{1.1}

\begin{tcolorbox}[rulebox, title={	\textbf{Rule 1:} \texttt{traffic light} $\wedge$ one of \{\texttt{bicycle}, \texttt{bus}, \texttt{train}\} \\}]
\begin{tabularx}{\linewidth}{
    >{\columncolor{green!12}}l
    >{\columncolor{green!12}}X
    >{\columncolor{red!10}}l
    >{\columncolor{red!10}}X
}
\toprule
\multicolumn{2}{>{\columncolor{green!12}}l}{\textit{Positive Variants}} &
\multicolumn{2}{>{\columncolor{red!10}}l}{\textit{Near Boundary Types}} \\
\midrule
\pitem{1}{\texttt{traffic light} $\wedge$ \texttt{bicycle}} & \nitem{1}{\texttt{bicycle} $\wedge$ $\neg$  \texttt{traffic light}} \\
\pitem{2}{\texttt{traffic light} $\wedge$ \texttt{bus}}     & \nitem{2}{\texttt{bus} $\wedge$ $\neg$  \texttt{traffic light}} \\
\pitem{3}{\texttt{traffic light} $\wedge$ \texttt{train}}   & \nitem{3}{\texttt{train} $\wedge$ $\neg$  \texttt{traffic light}} \\
 &                                                      & \nitem{4}{\texttt{traffic light} $\wedge$ none of \texttt{bicycle}/\texttt{bus}/\texttt{train}} \\
\bottomrule
\end{tabularx}
\end{tcolorbox}

\begin{tcolorbox}[rulebox, title={\textbf{Rule 2:} Exactly two categories of \{\texttt{bottle}, \texttt{cup}, \texttt{pizza}\}}]
\begin{tabularx}{\linewidth}{
    >{\columncolor{green!12}}l
    >{\columncolor{green!12}}X
    >{\columncolor{red!10}}l
    >{\columncolor{red!10}}X
}
\toprule
\multicolumn{2}{>{\columncolor{green!12}}l}{\textit{Positive Variants}} &
\multicolumn{2}{>{\columncolor{red!10}}l}{\textit{Near Boundary Types}} \\
\midrule
\pitem{1}{\texttt{bottle} $\wedge$ \texttt{cup} $\wedge$ $\neg$  \texttt{pizza}}   & \nitem{1}{\texttt{bottle} $\wedge$ $\neg$  \texttt{cup} $\wedge$ $\neg$  \texttt{pizza}} \\
\pitem{2}{\texttt{bottle} $\wedge$ \texttt{pizza} $\wedge$ $\neg$  \texttt{cup}} & \nitem{2}{\texttt{cup} $\wedge$ $\neg$  \texttt{bottle} $\wedge$ $\neg$  \texttt{pizza}} \\
\pitem{3}{\texttt{cup} $\wedge$ \texttt{pizza} $\wedge$ $\neg$  \texttt{bottle}}    & \nitem{3}{\texttt{pizza} $\wedge$ $\neg$  \texttt{bottle} $\wedge$ $\neg$  \texttt{cup}} \\
 &                                            & \nitem{4}{\texttt{bottle} $\wedge$ \texttt{cup} $\wedge$ \texttt{pizza}} \\
\bottomrule
\end{tabularx}
\end{tcolorbox}

\begin{tcolorbox}[rulebox, title={\scriptsize{\textbf{Rule 3:} At least two objects of the same category of \{\texttt{cow}, \texttt{elephant}, \texttt{sheep}\} $\wedge$ no \texttt{person}}}]
\begin{tabularx}{\linewidth}{
    >{\columncolor{green!12}}l
    >{\columncolor{green!12}}X
    >{\columncolor{red!10}}l
    >{\columncolor{red!10}}X
}
\toprule
\multicolumn{2}{>{\columncolor{green!12}}l}{\textit{Positive Variants}} &
\multicolumn{2}{>{\columncolor{red!10}}l}{\textit{Near Boundary Types}} \\
\midrule
\pitem{1}{Two or more \texttt{cow} $\wedge$ $\neg$ \texttt{person}}      & \nitem{1}{Exactly one \texttt{elephant} $\wedge$ $\neg$ \texttt{person}} \\
\pitem{2}{Two or more \texttt{elephant} $\wedge$ $\neg$ \texttt{person}} & \nitem{2}{Exactly one \texttt{cow} $\wedge$ $\neg$ \texttt{person}} \\
\pitem{3}{Two or more \texttt{sheep} $\wedge$ $\neg$ \texttt{person}}    & \nitem{3}{Exactly one \texttt{sheep} $\wedge$ $\neg$ \texttt{person}} \\
 &                                                              & \nitem{4}{Two or more \texttt{elephant} $\wedge$ \texttt{person}} \\
 &                                                              & \nitem{5}{Two or more \texttt{cow} $\wedge$ \texttt{person}} \\
 &                                                              & \nitem{6}{Two or more \texttt{sheep} $\wedge$ \texttt{person}} \\
\bottomrule
\end{tabularx}
\end{tcolorbox}

\begin{tcolorbox}[rulebox, title={\textbf{Rule 4:} Either \texttt{dog} or \texttt{car}}]
\begin{tabularx}{\linewidth}{
    >{\columncolor{green!12}}l
    >{\columncolor{green!12}}X
    >{\columncolor{red!10}}l
    >{\columncolor{red!10}}X
}
\toprule
\multicolumn{2}{>{\columncolor{green!12}}l}{\textit{Positive Variants}} &
\multicolumn{2}{>{\columncolor{red!10}}l}{\textit{Near Boundary Types}} \\
\midrule
\pitem{1}{\texttt{dog} $\wedge$ $\neg$ \texttt{car}} & \nitem{1}{\texttt{dog} $\wedge$ \texttt{car}} \\
\pitem{2}{\texttt{car} $\wedge$ $\neg$ \texttt{dog}} & \nitem{2}{$\neg$ \texttt{dog} $\wedge$ $\neg$ \texttt{car}} \\
\bottomrule
\end{tabularx}
\end{tcolorbox}

\begin{tcolorbox}[rulebox, title={\textbf{Rule 5:} Exactly three \texttt{bowl} $\lor$ exactly three \texttt{cup}}]
\begin{tabularx}{\linewidth}{
    >{\columncolor{green!12}}l
    >{\columncolor{green!12}}X
    >{\columncolor{red!10}}l
    >{\columncolor{red!10}}X
}
\toprule
\multicolumn{2}{>{\columncolor{green!12}}l}{\textit{Positive Variants}} &
\multicolumn{2}{>{\columncolor{red!10}}l}{\textit{Near Boundary Types}} \\
\midrule
\pitem{1}{Exactly 3 \texttt{bowl}} & \nitem{1}{1--2 \texttt{cup}} \\
\pitem{2}{Exactly 3 \texttt{cup}}  & \nitem{2}{4 or more \texttt{cup}} \\
 &                                 & \nitem{3}{1--2 \texttt{bowl}} \\
 &                                 & \nitem{4}{4 or more \texttt{bowl}} \\
\bottomrule
\end{tabularx}
\end{tcolorbox}

\begin{tcolorbox}[rulebox, title={\textbf{Rule 6:} More \texttt{car} than \texttt{truck} $\wedge$ at least one of each}]
\begin{tabularx}{\linewidth}{
    >{\columncolor{green!12}}l
    >{\columncolor{green!12}}X
    >{\columncolor{red!10}}l
    >{\columncolor{red!10}}X
}
\toprule
\multicolumn{2}{>{\columncolor{green!12}}l}{\textit{Positive Variants}} &
\multicolumn{2}{>{\columncolor{red!10}}l}{\textit{Near Boundary Types}} \\
\midrule
\pitem{1}{More \texttt{car} than \texttt{truck}; both present} & \nitem{1}{\texttt{car} $\wedge$ $\neg$ \texttt{truck}} \\
 &                                                             & \nitem{2}{Equal number of \texttt{car} and \texttt{truck}; both present} \\
 &                                                             & \nitem{3}{More \texttt{truck} than \texttt{car}; both present} \\
\bottomrule
\end{tabularx}
\end{tcolorbox}

\begin{tcolorbox}[rulebox, title={\textbf{Rule 7:} (\texttt{couch} $\lor$ \texttt{chair}) $\wedge$ $\neg$ \texttt{person}}]
\begin{tabularx}{\linewidth}{
    >{\columncolor{green!12}}l
    >{\columncolor{green!12}}X
    >{\columncolor{red!10}}l
    >{\columncolor{red!10}}X
}
\toprule
\multicolumn{2}{>{\columncolor{green!12}}l}{\textit{Positive Variants}} &
\multicolumn{2}{>{\columncolor{red!10}}l}{\textit{Near Boundary Types}} \\
\midrule
\pitem{1}{\texttt{couch} $\wedge$ $\neg$ \texttt{person}} & \nitem{1}{ $\wedge$ \texttt{person}} \\
\pitem{2}{\texttt{chair} $\wedge$ $\neg$ \texttt{person}} & \nitem{2}{\texttt{chair} $\wedge$ \texttt{person}} \\
& & \nitem{3}{$\neg$ \texttt{couch} $\wedge$ $\neg$ \texttt{chair} $\wedge$ $\neg$ \texttt{person}} \\
\bottomrule
\end{tabularx}
\end{tcolorbox}

\begin{tcolorbox}[rulebox, title={\textbf{Rule 8:} Exactly one category of \{\texttt{bicycle, motorcycle, car, bus}\}}]
\begin{tabularx}{\linewidth}{
    >{\columncolor{green!12}}l
    >{\columncolor{green!12}}X
    >{\columncolor{red!10}}l
    >{\columncolor{red!10}}X
}
\toprule
\multicolumn{2}{>{\columncolor{green!12}}l}{\textit{Positive Variants}} &
\multicolumn{2}{>{\columncolor{red!10}}l}{\textit{Near Boundary Types}} \\
\midrule
\pitem{1}{\texttt{bicycle} $\wedge$ none of \texttt{motorcycle}/\texttt{car}/\texttt{bus}}    & \nitem{1}{\texttt{bicycle} $\wedge$ \texttt{motorcycle} $\wedge$ none of \texttt{car}/\texttt{bus}} \\
\pitem{2}{\texttt{motorcycle} $\wedge$ none of \texttt{bicycle}/\texttt{car}/\texttt{bus}}    & \nitem{2}{\texttt{bicycle} $\wedge$ \texttt{car} $\wedge$ none of \texttt{motorcycle}/\texttt{bus}} \\
\pitem{3}{\texttt{car} $\wedge$ none of \texttt{bicycle}/\texttt{motorcycle}/\texttt{bus}}    & \nitem{3}{\texttt{bicycle} $\wedge$ \texttt{bus} $\wedge$ none of \texttt{motorcycle}/\texttt{car}} \\
\pitem{4}{\texttt{bus} $\wedge$ none of \texttt{bicycle}/\texttt{motorcycle}/\texttt{car}}    & \nitem{4}{\texttt{motorcycle} $\wedge$ \texttt{car} $\wedge$ none of \texttt{bicycle}/\texttt{bus}} \\
 &                                                                                         & \nitem{5}{\texttt{motorcycle} $\wedge$ \texttt{bus} $\wedge$ none of \texttt{bicycle}/\texttt{car}} \\
 &                                                                                         & \nitem{6}{\texttt{car} $\wedge$ \texttt{bus} $\wedge$ none of \texttt{bicycle}/\texttt{motorcycle}} \\
\bottomrule
\end{tabularx}
\end{tcolorbox}

\begin{tcolorbox}[rulebox, title={\textbf{Rule 9:} \texttt{person} $\wedge$ (Either \texttt{bicycle} or \texttt{car})}]
\begin{tabularx}{\linewidth}{
    >{\columncolor{green!12}}l
    >{\columncolor{green!12}}X
    >{\columncolor{red!10}}l
    >{\columncolor{red!10}}X
}
\toprule
\multicolumn{2}{>{\columncolor{green!12}}l}{\textit{Positive Variants}} &
\multicolumn{2}{>{\columncolor{red!10}}l}{\textit{Near Boundary Types}} \\
\midrule
\pitem{1}{\texttt{person} $\wedge$ \texttt{bicycle} $\wedge$ $\neg$ \texttt{car}} & \nitem{1}{\texttt{person} $\wedge$ \texttt{bicycle} $\wedge$ \texttt{car}} \\
\pitem{2}{\texttt{person} $\wedge$ \texttt{car} $\wedge$ $\neg$ \texttt{bicycle}} & \nitem{2}{\texttt{person} $\wedge$ $\neg$ \texttt{bicycle} $\wedge$ $\neg$ \texttt{car}} \\
 &                                                                   & \nitem{3}{\texttt{bicycle} $\wedge$ $\neg$ \texttt{car} $\wedge$ $\neg$ \texttt{person}} \\
 &                                                                   & \nitem{4}{\texttt{car} $\wedge$ $\neg$ \texttt{bicycle} $\wedge$ $\neg$ \texttt{person}} \\
\bottomrule
\end{tabularx}
\end{tcolorbox}

\begin{tcolorbox}[rulebox, title={\textbf{Rule 10:} Exactly as many \texttt{person} as \texttt{surfboard}}]
\begin{tabularx}{\linewidth}{
    >{\columncolor{green!12}}l
    >{\columncolor{green!12}}X
    >{\columncolor{red!10}}l
    >{\columncolor{red!10}}X
}
\toprule
\multicolumn{2}{>{\columncolor{green!12}}l}{\textit{Positive Variants}} &
\multicolumn{2}{>{\columncolor{red!10}}l}{\textit{Near Boundary Types}} \\
\midrule
\pitem{1}{Equal number of \texttt{person} and \texttt{surfboard}; both present} & \nitem{1}{\texttt{surfboard} $\wedge$ $\neg$ \texttt{person}} \\
 &                                                                               & \nitem{2}{\texttt{person} $\wedge$ $\neg$ \texttt{surfboard}} \\
 &                                                                               & \nitem{3}{More \texttt{surfboard} than \texttt{person}; both present} \\
 &                                                                               & \nitem{4}{More \texttt{person} than \texttt{surfboard}; both present} \\
\bottomrule
\end{tabularx}
\end{tcolorbox}

\subsection{Number of Samples per Variant and NB Type}
\label{app:sampling_details}
In the tables \autoref{tab:cocologicv2-counts-full} and \autoref{tab:cocologicv2-counts-fs}, the exact number of samples for each positive variant and NB type is presented. It is noteworthy that, due to the small size of the COCO validation set, the total number of samples in the \data test set is low for some NB types. 

\begin{table}[ht]
\centering
\caption{\data per-rule sample counts. P$i$ and N$i$ denote positive variants and NB types for the respective rules. All other samples that are not a positive variant or an NB type are considered FB samples.}
\label{tab:cocologicv2-counts-full}
\resizebox{\linewidth}{!}{
\begin{tabular}{l l c c c c c c c c c c}
\toprule
Rule & Split & \multicolumn{4}{c}{Positive Variants} & \multicolumn{6}{c}{Near Boundary} \\
\cmidrule(lr){3-6} \cmidrule(lr){7-12}
 &  & PV1 & PV2 & PV3 & PV4 & NB1 & NB2 & NB3 & NB4 & NB5 & NB6 \\
\midrule
\multirow{2}{*}{Rule 1 (Signal and Ride)} & train & 344 & 734 & 376 & -- & 1926 & 1227 & 541 & 616 & -- & -- \\
 & test & 24 & 35 & 23 & -- & 125 & 154 & 134 & 117 & -- & -- \\
\midrule
\multirow{2}{*}{Rule 2 (Double Serving)} & train & 1106 & 311 & 657 & -- & 639 & 954 & 521 & 351 & -- & -- \\
 & test & 112 & 20 & 39 & -- & 231 & 223 & 78 & 16 & -- & -- \\
\midrule
\multirow{2}{*}{Rule 3 (Herd Alone)} & train & 936 & 801 & 948 & -- & 476 & 370 & 167 & 368 & 336 & 275 \\
 & test & 48 & 42 & 31 & -- & 13 & 14 & 8 & 15 & 11 & 15 \\
\midrule
\multirow{2}{*}{Rule 4 (Either Dog or Car)} & train & 1308 & 3308 & -- & -- & 481 & 113190 & -- & -- & -- & -- \\
 & test & 158 & 516 & -- & -- & 19 & 4307 & -- & -- & -- & -- \\
\midrule
\multirow{2}{*}{Rule 5 (Three of a Kind)} & train & 608 & 842 & -- & -- & 1632 & 594 & 957 & 500 & -- & -- \\
 & test & 25 & 29 & -- & -- & 296 & 65 & 256 & 33 & -- & -- \\
\midrule
\multirow{2}{*}{Rule 6 (Car Majority)} & train & 1213 & -- & -- & -- & 1247 & 523 & 560 & -- & -- & -- \\
 & test & 104 & -- & -- & -- & 260 & 37 & 109 & -- & -- & -- \\
\midrule
\multirow{2}{*}{Rule 7 (Empty Seat)} & train & 1102 & 1338 & -- & -- & 500 & 1442 & 9697 & -- & -- & -- \\
 & test & 81 & 189 & -- & -- & 90 & 367 & 1344 & -- & -- & -- \\
\midrule
\multirow{2}{*}{Rule 8 (Single Mode Traffic)} & train & 1334 & 1071 & 1991 & 1007 & 205 & 500 & 137 & 500 & 70 & 639 \\
 & test & 75 & 81 & 356 & 76 & 14 & 33 & 6 & 36 & 3 & 78 \\
\midrule
\multirow{2}{*}{Rule 9 (Personal Transport)} & train & 1205 & 2201 & -- & -- & 515 & 9080 & 475 & 998 & -- & -- \\
 & test & 77 & 314 & -- & -- & 45 & 1488 & 20 & 169 & -- & -- \\
\midrule
\multirow{2}{*}{Rule 10 (Surf Trip)} & train & 1088 & -- & -- & -- & 38 & 7796 & 133 & 500 & -- & -- \\
 & test & 97 & -- & -- & -- & 3 & 1175 & 3 & 42 & -- & -- \\
\bottomrule
\end{tabular}
}
\end{table}

\begin{table}[ht]
\centering
\caption{\dataf per-rule sample counts. P$i$ and N$i$ denote positive variants and NB types for the respective rules. All other samples that are not a positive variant or an NB type are considered FB samples.}
\label{tab:cocologicv2-counts-fs}
\resizebox{\linewidth}{!}{
\begin{tabular}{l l c c c c c c c c c c}
\toprule
Rule & Split & \multicolumn{4}{c}{Positive Variants} & \multicolumn{6}{c}{Near Boundary} \\
\cmidrule(lr){3-6} \cmidrule(lr){7-12}
 &  & PV1 & PV2 & PV3 & PV4 & NB1 & NB2 & NB3 & NB4 & NB5 & NB6 \\
\midrule
\multirow{2}{*}{Rule 1 (Signal and Ride)} & train & 3 & 3 & 2 & -- & 3 & 3 & 3 & 3 & -- & -- \\
 & test & 7 & 7 & 6 & -- & 4 & 4 & 3 & 4 & -- & -- \\
\midrule
\multirow{2}{*}{Rule 2 (Double Serving)} & train & 3 & 3 & 2 & -- & 3 & 3 & 3 & 3 & -- & -- \\
 & test & 7 & 7 & 6 & -- & 4 & 4 & 4 & 3 & -- & -- \\
\midrule
\multirow{2}{*}{Rule 3 (Herd Alone)} & train & 3 & 3 & 2 & -- & 2 & 2 & 2 & 2 & 2 & 2 \\
 & test & 7 & 7 & 6 & -- & 3 & 3 & 3 & 2 & 2 & 3 \\
\midrule
\multirow{2}{*}{Rule 4 (Either Dog or Car)} & train & 4 & 4 & -- & -- & 8 & 8 & -- & -- & -- & -- \\
 & test & 10 & 10 & -- & -- & 10 & 10 & -- & -- & -- & -- \\
\midrule
\multirow{2}{*}{Rule 5 (Three of a Kind)} & train & 4 & 4 & -- & -- & 3 & 3 & 3 & 3 & -- & -- \\
 & test & 10 & 10 & -- & -- & 4 & 3 & 4 & 4 & -- & -- \\
\midrule
\multirow{2}{*}{Rule 6 (Car Majority)} & train & 8 & -- & -- & -- & 4 & 4 & 4 & -- & -- & -- \\
 & test & 20 & -- & -- & -- & 5 & 4 & 5 & -- & -- & -- \\
\midrule
\multirow{2}{*}{Rule 7 (Empty Seat)} & train & 4 & 4 & -- & -- & 4 & 4 & 3 & -- & -- & -- \\
 & test & 10 & 10 & -- & -- & 5 & 5 & 4 & -- & -- & -- \\
\midrule
\multirow{2}{*}{Rule 8 (Single Mode Traffic)} & train & 2 & 2 & 2 & 2 & 2 & 2 & 2 & 2 & 2 & 2 \\
 & test & 5 & 5 & 5 & 5 & 3 & 3 & 2 & 3 & 2 & 3 \\
\midrule
\multirow{2}{*}{Rule 9 (Personal Transport)} & train & 4 & 4 & -- & -- & 3 & 3 & 3 & 3 & -- & -- \\
 & test & 10 & 10 & -- & -- & 4 & 4 & 3 & 4 & -- & -- \\
\midrule
\multirow{2}{*}{Rule 10 (Surf Trip)} & train & 8 & -- & -- & -- & 3 & 3 & 3 & 3 & -- & -- \\
 & test & 20 & -- & -- & -- & 3 & 4 & 2 & 4 & -- & -- \\
\bottomrule
\end{tabular}
}
\end{table}

\FloatBarrier

\subsection{Detailed Results}\label{app:result_details}

\begin{figure}
    \centering
    \includegraphics[width=\linewidth]{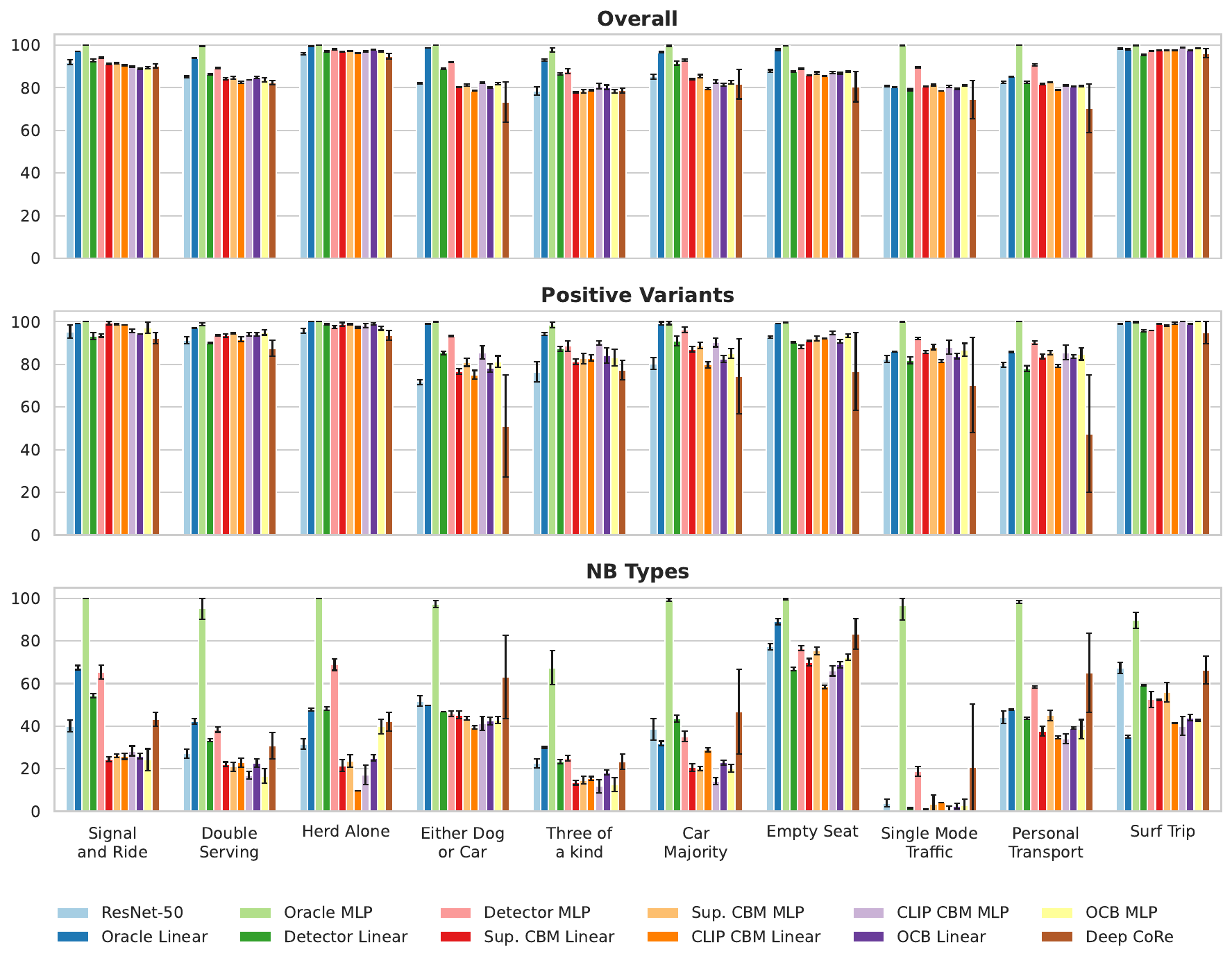}
    \caption{Per-rule performance on \data. Overall balanced accuracy (top), accuracy averaged over all positive variants of a rule (middle), and accuracy averaged over all NB types of a rule (bottom) on the full dataset.}
    \label{fig:full_results_combined}
\end{figure}

In \autoref{fig:full_results_combined}, the performance of all models on each individual rule is shown. The results show that the overall performance does not correlate much with performance on the NB types.
Overall, models perform best on Surf Trip and Herd Alone, but the NB performance stays around random guessing.
The rule with the best NB performance is Empty Seat, which is reasonable as the NB types are quite simple and can be separated from positive samples by identifying a person.
When considering rules that require object counting (i.e., Herd Alone, Three of a Kind, Car Majority, and Surf Trip), detector-based CBMs and OCB do not perform much better despite having access to object counts directly.
The noise in the concept encoder makes the object counts insufficient to gain a substantial advantage on these rules. 

\subsection{Experimental Details} \label{app:experimental_details}
\textbf{Code.} The code will be made public upon acceptance.
It also includes detailed information on hyperparameters and experimental configurations. 
In the following, we provide an overview of the experimental setup.

\textbf{Training Models on \data.} All models were trained on the training split of \data, where 10\% of the samples were randomly used as a validation set. 
Learning rate, number of epochs trained, and model-specific parameters were optimized using the validation set. 
The final models were trained for 5 seeds with the best parameter configuration and early stopping. 
For the CLIP-based CBM and OCB, the textual vocabulary is the list of all object-category names of MSCOCO without further modifications. 
For DCR, the encoder is a concept embedding model \cite{espinosa2022concept} (also instantiated with a pretrained ResNet-50) that is jointly trained with the logic-based predictor as in \cite{barbiero2023interpretable}.

\textbf{Training Models on \dataf.} 
The training setup differs from the full version due to the small number of training instances and the different general setup. 
For this version, one model is trained to predict each rule only on the training samples that belong to this rule, i.e., 10 models are trained on 24 samples each, and each is evaluated only on the 40 test samples for its rule.
As the training set for each of these models contains only 24 samples, using a dedicated validation set is of little use.
Instead, the hyperparameters of all models were optimized using leave-one-out cross-validation. 
Besides these differences, the models were trained similarly to the full version, with the same hyperparameters being optimized. 
The concept encoders were fixed for all CBMs except DCR, which jointly trains the encoder and the predictor. 
The ResNet-50 model was also fine-tuned directly on \dataf.

\textbf{In-Context Learning on \dataf.}
For the In-Context Learning baseline, we provide the models with the few-shot images of a task and ask them to infer the underlying rule. All models are run with their default reasoning effort, and \texttt{max\_tokens} is set to 
$32768$ to allow for long reasoning traces. At test time, the model is then shown each test image individually, together with the inferred rule, and asked to classify the image according to that rule. Each model is evaluated with 5 seeds.

\begin{promptbox}{Prompt for In-Context Learning}
You are given {n} images. Each image depicts a scene with specific objects, interactions, and environments.
Your task is to determine the underlying concept that distinguishes the positive examples from the negative examples, based on the objects, their properties, and the actions occurring in each scene.

- The first {m} images are positive examples.
- The remaining {o} images are negative examples.

## Task

Identify the rule that defines the positive examples:
- The rule must apply to all positive examples.
- The rule must not apply to any negative example.

## Step-by-Step Process

1. Image Analysis:
- Carefully describe each image, noting objects, their attributes, and conceptual features (such as relationships, actions, or settings).

2. Rule Derivation:
- From your analysis, infer the rule that uniquely characterizes the positive examples.
- Confirm that the rule does not hold for the negative examples.

## Final Answer Format

Provide your final answer in the following format:

```python
answer = {
'rule': '[RULE]',
}
```

Ensure that the rule is clearly defined and concise.
\end{promptbox}

\begin{promptbox}{Test Prompt}
Given the rule '{response}', determine if the image follows the rule or not. Answer with 'Yes' or 'No', nothing else.
\end{promptbox}

\textbf{Program Synthesis on \dataf.} 
For the Vision-Language Program approach, we use the DSL shown in \autoref{tab:dsl}. It comprises the VLM function \texttt{get\_objects}, logical functions that check for the existence of objects and properties, as well as counting functions, arithmetic functions, and Boolean logic operators, all of which were already used in the original paper. We extend the DSL with a new primitive, \texttt{addb}, which adds an integer and a Boolean, enabling us to count how many of several logical statements are true. Programs are enumerated via heap search with a budget of 300 seconds and a maximum program depth of 8.

\begin{table}[t]
  \centering
    \caption{Primitives of the DSL used for VLP and their types. Arrows associate to the right, i.e.\ $A \to B \to C$ denotes a curried function $A \to (B \to C)$. $[[\textsc{Str}]]$ abbreviates a list of lists of strings.}
  \label{tab:dsl}
  \begin{tabular}{ll}
    \toprule
    \textbf{Primitive} & \textbf{Type} \\
    \midrule
    \multicolumn{2}{l}{\emph{Perception}} \\
    \texttt{get\_objects} & $\textsc{Img} \to [[\textsc{Str}]]$ \\
    \midrule
    \multicolumn{2}{l}{\emph{Existence predicates}} \\
    \texttt{exists\_object} & $[[\textsc{Str}]] \to \textsc{Obj} \to \textsc{Bool}$ \\
    \texttt{exists\_object\_with\_property} & $[[\textsc{Str}]] \to \textsc{Obj} \to \textsc{Prop} \to \textsc{Bool}$ \\
    \texttt{exists\_property} & $[[\textsc{Str}]] \to \textsc{Prop} \to \textsc{Bool}$ \\
    \midrule
    \multicolumn{2}{l}{\emph{Counting}} \\
    \texttt{count\_object\_in\_img} & $[[\textsc{Str}]] \to \textsc{Obj} \to \textsc{Int}$ \\
    \texttt{count\_all\_objects} & $[[\textsc{Str}]] \to \textsc{Int}$ \\
    \midrule
    \multicolumn{2}{l}{\emph{Boolean logic}} \\
    \texttt{and}, \texttt{or}, \texttt{xor} & $\textsc{Bool} \to \textsc{Bool} \to \textsc{Bool}$ \\
    \texttt{not} & $\textsc{Bool} \to \textsc{Bool}$ \\
    \midrule
    \multicolumn{2}{l}{\emph{Arithmetic and comparison}} \\
    \texttt{gt?}, \texttt{eq?} & $\textsc{Int} \to \textsc{Int} \to \textsc{Bool}$ \\
    \texttt{addb} & $\textsc{Int} \to \textsc{Bool} \to \textsc{Int}$ \\
    \texttt{0}, \texttt{1}, \texttt{2}, \texttt{3} & $\textsc{Int}$ \\
    \bottomrule
  \end{tabular}
\end{table}

\end{document}